\documentclass[11pt,a4paper]{article}
\usepackage[hyperref]{acl2019}
\usepackage{times}
\usepackage{latexsym}

\usepackage{url}

\usepackage{subcaption}

\usepackage{tikz}
\usetikzlibrary{calc}
\usepackage{pgfplots}
\pgfplotsset{compat=1.12}

\definecolor{myred}{cmyk}{0.0000,0.4612,0.2082,0.0392}
\definecolor{mygreen}{cmyk}{0.3917,0.0000,0.3500,0.0588}
\definecolor{myorange}{cmyk}{0,0.33,0.81,0}

\usepackage{lingmacros}

\usepackage[english]{babel}
\usepackage[utf8]{inputenc}
\usepackage{tikzsymbols}
\usepackage{textcomp}
\newcommand{\perfect}{\Smiley[][green]}

\newcommand{\medium}{\Neutrey[][myorange]}

\usepackage{amstext}

\usepackage{booktabs}
\usepackage{multicol}

\usepackage{enumitem}

\usepackage{graphicx}
\usepackage{MnSymbol,wasysym}
\usepackage{mdframed}

\usepackage{tikz}
\usepackage{collcell}
\usepackage{colortbl}

\newcommand*{\MinNumber}{-11}%
\newcommand*{\MidNumber}{0} %
\newcommand*{\MaxNumber}{11}%
\newcommand{\ApplyGradient}[1]{%
        \ifdim #1 pt > \MidNumber pt
            \pgfmathsetmacro{\PercentColor}{max(min(100.0*(#1 - \MidNumber)/(\MaxNumber-\MidNumber),100.0),0.00)} %
            \xdef\PercentColor{\PercentColor}%
            \cellcolor{mygreen!\PercentColor!white}{#1}
        \else
            \pgfmathsetmacro{\PercentColor}{max(min(100.0*(\MidNumber - #1)/(\MidNumber-\MinNumber),100.0),0.00)} %
            \xdef\PercentColor{\PercentColor}%
            \cellcolor{myred!\PercentColor!white}{#1}
        \fi
}
\newcolumntype{R}{>{\collectcell\ApplyGradient}c<{\endcollectcell}}
\aclfinalcopy 

\title{DiaBLa: A Corpus of Bilingual Spontaneous Written Dialogues \\for Machine Translation}

\author{Rachel Bawden$^1$\thanks{\hspace{2mm}This work was started while the first author was
  a PhD student at LIMSI, CNRS, Univ.~Paris-Sud, Université Paris-Saclay.} \qquad Sophie Rosset$^2$ \qquad Thomas Lavergne$^{2,3}$ \qquad Eric  Bilinski$^2$\\
$^1$School of Informatics, University of Edinburgh, Scotland \\
$^2$LIMSI, CNRS, Université Paris-Saclay, Orsay, France \\
$^3$Univ.~Paris-Sud, Orsay, France\\
\texttt {rachel.bawden@ed.ac.uk \quad lastname@limsi.fr}
}

\date{}

\begin{document}
\maketitle
\begin{abstract}
  We present a new English-French test set for the evaluation of Machine Translation (MT) for
  informal, written bilingual dialogue. The test set contains 144 spontaneous dialogues (5,700+
  sentences) between native English and French speakers, mediated by one of two neural MT
  systems in a range of role-play settings.  The dialogues are accompanied by fine-grained
  sentence-level judgments of MT quality, produced by the dialogue participants themselves, as well
  as by manually normalised versions and reference translations produced \textit{a
    posteriori}.  The motivation for the corpus is two-fold:~to provide (i)~a unique resource for
  evaluating MT models, and (ii)~a corpus for the analysis of MT-mediated communication. We provide
  a preliminary analysis of the corpus to confirm that the participants' judgments reveal
  perceptible differences in MT quality between the two MT systems used.
\end{abstract}

 \newcommand{\nbParticipants}{75}
 \newcommand{\lastParticipantId}{74}
 \newcommand{\nbParticipantsEn}{37}
 \newcommand{\nbParticipantsFr}{38}
 \newcommand{\totalNbDialogues}{144}
 \newcommand{\totalNbSentences}{5,748}
 \newcommand{\meanNbTurns}{15.0}
 \newcommand{\percentDialoguesMoreThanTen}{100.0}
 \newcommand{\percentDialoguesMoreThanTwenty}{97.9}
 \newcommand{\percentDialoguesMoreThanThirty}{95.1}
 \newcommand{\percentDialoguesMoreThanThirtyfive}{75.7}
 \newcommand{\percentDialoguesMoreThanForty}{47.9}
 \newcommand{\percentDialoguesMoreThanFifty}{9.7}
 \newcommand{\percentDialoguesMoreThanSixty}{2.1}
 \newcommand{\meanNbOrigToksPerSent}{10.0}
 \newcommand{\meanNbOrigToksPerSentFR}{10.2}
 \newcommand{\meanNbOrigToksPerSentEN}{9.8}
 \newcommand{\meanNbTransToksPerSentFR}{10.1}
 \newcommand{\meanNbTransToksPerSentEN}{9.6}
 \newcommand{\baselineTuTotal}{240}
 \newcommand{\baselineVousTotal}{174}
 \newcommand{\contexutalTuTotal}{269}
 \newcommand{\contexutalVousTotal}{148}
 \newcommand{\baselineTuTu}{49}
 \newcommand{\baselineVousTu}{31}
 \newcommand{\contexutalTuTu}{55}
 \newcommand{\contexutalVousTu}{23}
 \newcommand{\baselineTuVous}{27}
 \newcommand{\baselineVousVous}{27}
 \newcommand{\contexutalTuVous}{22}
 \newcommand{\contexutalVousVous}{28}
 \newcommand{\percentDiffTuTu}{9}
 \newcommand{\percentDiffVousVous}{6}

\section{Introduction}

The use of Machine Translation (MT) to translate everyday, written exchanges is becoming
increasingly commonplace; translation tools now regularly appear on chat applications and 
social networking sites to enable cross-lingual communciation. MT systems must therefore be
able to deal with a wide variety of topics, styles and vocabulary. Importantly, the
translation of dialogue requires translating sentences coherently with respect to the
conversational flow in order for all aspects of the exchange, including speaker intent, attitude and
style, to be correctly communicated. 




It is important to have realistic data to evaluate MT models and to guide future MT research for
informal, written exchanges. In this article, we present DiaBLa (\textit{Dialogue BiLingue}
`Bilingual Dialogue'), a new test set of English-French spontaneous written dialogues
mediated by MT,\footnote{Participants write and receive messages in their respective native language
  thanks to MT systems translating between the two languages.}  obtained by crowdsourcing and
covering a range of dialogue topics and annotated with fine-grained human judgments of MT quality.
To our knowledge, this is the first corpus of its kind. Our data collection protocol is designed to
encourage spontaneous dialogue between speakers of two languages, using role-play scenarios to
provide conversation material. Sentence-level human judgments of the quality of the MT systems are
provided by the participants themselves while they are actively engaged in dialogue. The result is a
rich bilingual test corpus of \totalNbDialogues{} dialogues, which are annotated with sentence-level MT quality
evaluations and human reference translations.

We discuss the potential of the corpus and the collection method for MT research in
Sec.~\ref{sec:potential}, both for MT evaluation and for the study of language behaviour
in informal dialogues. In Sec.~\ref{sec:protocol} we describe the data collection
protocol and interface. We describe basic characteristics and examples of the corpus in
Sec.~\ref{sec:corpus}. This includes a description of the annotation layers (normalised
versions, human reference translations and human MT quality judgments). We illustrate the
usefulness of the human evaluation by providing a comparison and analysis of the MT systems used
(Sec.~\ref{sec:eval}). We compare two different types of MT system, a baseline model and a mildly
context-aware model. Finally, we provide plans for future work on the corpus in Sec.~\ref{sec:future}.
The corpus, interface, scripts and participation guidelines are freely available under a CC
BY-SA 3.0 licence.\footnote{\url{https://github.com/rbawden/DiaBLa-dataset}}

\subsection{Related work}

A number of corpora of informal data do exist. However they either cover different domains or are
not designed with the same aim in mind. \mbox{OpenSubtitles2016}
\citep{lison_opensubtitles2016:_2016} is a large-scale corpus of film subtitles, from a variety of
domains, making for very heterogeneous content. However the conversations are scripted rather than
being spontaneous, and are translations of monolingual texts, rather than being bilingual
conversations. The MSLT corpus \citep{federmann_microsoft_2016} is designed as a bilingual corpus,
and is based on oral dialogues produced by bilingual speakers, who understand the other speaker's
original utterances. This means that it is not possible to analyse the impact that using MT has on
the interaction between participants. Other bilingual task-orientated corpora exist, for example
BTEC (Basic Travel Expression Corpus) \citep{takezawa_toward_2002}, SLDB (Spoken Language DataBase)
\citep{morimoto_speech_1994} and Field Experiment Data of \citet{Takezawa_multilingual_2007}, which
is the most similar corpus to our own in that it contains MT-mediated dialogue. However these
corpora are restricted to the travel/hotellery domains, therefore not allowing the same variety of
conversation topic as our corpus. Human judgments for the overall quality are provided for the third
corpus, but only at a very coarse-grained level. Feedback about the participants' perception of MT
quality is therefore of a limited nature in terms of MT evaluation; sentence-level evaluations are
not provided.


\section{Motivation}\label{sec:potential}

The main aim of our corpus is as a test set to evaluate MT models in an informal setting in which
communication is mediated by MT systems. However, the corpus can also be of interest
for studying the type of language used in the dialogues as well as the way in which human
interaction is affected by use of MT as a mediation tool. We develop these two motivations here,
starting with the corpus' utility for MT evaluation (Sec.~\ref{sec:mt-eval}) and then discussing
the corpus' potential for analysis of MT-assisted interaction (Sec.~\ref{sec:mt-interaction}).

\subsection{MT evaluation}\label{sec:mt-eval}
The corpus is useful for MT evaluation in three ways: as (i)~a test set for 
automatically evaluating new models, (ii)~a challenge set for manual evaluation, and (iii)~a
validation of the effectiveness of the protocol to collect new dialogues and to
compare new translation models in the future.

\paragraph{Test set for automatic evaluation}
The test set provides an example of spontaneously produced utterances in a real, unscripted
setting. It could be particularly useful for evaluating contextual MT models due to the dialogic
nature of the utterances, the need to take into account previous MT outputs and the presence of
metadata. 

\paragraph{Challenge set for manual evaluation}
It can be used as a basis for manual evaluation (as a challenge set). The sentence-level
human judgments provided can be used as an indicator as to which sentences were the
most challenging for MT. Manual evaluation of new translations of our test set can then be
guided towards those sentences whose translations were marked as \textit{poor}, to provide an
informed idea of the quality of the new models of these difficult examples, and to encourage
development for particularly challenging phenomena.

\paragraph{Validation of the protocol for the collection of human judgments of MT quality}
Human evaluation remains the most accurate form of MT evaluation, especially for understanding which
aspects of language pose difficulties for translation. While hand-crafted examples and challenge
sets provide the means to test particular phenomena \citep{king_using_1990,isabelle_challenge_2017},
it is also important to observe and evaluate the quality of translation on real, spontaneously
produced texts. Our corpus provides this opportunity, as it contains spontaneous productions by
human participants and is richly annotated for MT quality by its end users (see
Sec.~\ref{sec:eval}). We provide a preliminary comparative evaluation of the two MT systems, in
order to show the utility of the human judgments collected. The same collection method can be
applied to new MT models for a similar evaluation.


\subsection{MT-assisted interaction}\label{sec:mt-interaction}
As MT systems are becoming more common online, it is important for them to take into account the
type of language that may be used and the way in which user behaviour may affect the system's
translation quality. Non-canonical syntactic structures, spelling and typing errors, text mimicking
speech, including pauses and reformulations, must be taken into account if MT systems are to be
used for successful communication in more informal environments. The language used in our corpus is
relatively clean in terms of spelling. However participants are encouraged to be natural with their
language, and therefore a fruitful direction would be in the analysis of the type of language used.

Another interesting aspect of human-MT interaction would be to study how users themselves adapt to
using such a tool during the dialogues. How do they deal with translation errors, particularly those
that make the dialogue incoherent? Do they adjust their language over time, and how do they indicate
when they have not understood correctly? An interesting line of research would be to use the
corpus to study users' communication strategies, for example by studying breakdowns
in communication as in \citep{Higashinaka_dialogue_2016}.

\section{Data collection and protocol}\label{sec:protocol}

We collected the dialogues via a dedicated web interface, 
allowing participants to register, log on
and chat. Each dialogue involves two speakers, a native French speaker and a native English
speaker. Each writes in their native language and the dialogue is mediated by two
MT systems, one translating French utterances into English and the
other translating English utterances into French.

\paragraph{Participants}
Participants are volunteers recruited by word of mouth and social media. They participated free of
charge, motivated by the fun of taking part in fictional role-play.
Users provide basic information: age bracket, gender, English and
French language ability, other languages spoken and whether they work in research or Natural
Language Processing (see Tab.~\ref{tab:users} for basic statistics).

\paragraph{Scenarios}
To provide inspiration, a role-play scenario is given at the start of each dialogue and roles are
randomly assigned to the speakers. We designed twelve different scenarios
(cf.~App.~\ref{A:scenarios}) two of which are shown in Fig.~\ref{fig:scenarios}.  The first
turn is also assigned randomly to one of the speakers to get the dialogue started. This information
is indicated at the top of the dialogue screen in the participants' native languages. A minimum number
of 15 sentences per speaker is recommended, and participants are informed once this threshold is
reached. Participants are told to play fictional characters and not to use any personal details. We
nevertheless anonymise the corpus prior to distribution to remove usernames mentioned in the text.

\begin{table}[h!]
\centering
\small

    \begin{tabular}{lrrr}
    \toprule
     & EN & FR & All \\
    \midrule
    Total number  & 37 & 38 & 75 \\ 
\quad\#Researchers & 7 & 17 & 24 \\ 
\quad\#Experience in NLP & 6 & 14 & 20 \\ 
Min. age & 18-24 & 18-24 & 18-24 \\ 
Max. age & 65-74 & 65-74 & 65-74 \\ 
Median age & 55-64 & 25-34 & 35-44 \\ 
Modal age & 55-64 & 25-34 & 25-34 \\ 

    \bottomrule
    \end{tabular}
    
\caption{\label{tab:users} Some characteristics of the participants.}
\end{table}

\begin{figure}[t!]
\begin{mdframed}
\small
\textbf{You are both stuck in a lift at work}\\
(1)~You are an employee and are with your boss.\\
(2)~You are the boss and are with an employee.\\

\textbf{You are in a retirement home.}\\
(1)~You are visiting and talking to an old friend.\\
(2)~You are a resident and you are talking with an old friend who is visiting you.
\end{mdframed}
\caption{\label{fig:scenarios} Two scenarios and their speaker roles.}
\end{figure}

\paragraph{Evaluation method}

The participants evaluate each other's translated sentences from a monolingual point of
view. The choice to use the participants to provide the MT evaluation is an important
part of our protocol: we can collect judgments on the fly, facilitating the evaluation
process, and it importantly means that the evaluation is performed from the point of view of
participants actively engaged in dialogue. Athough some errors may go unnoticed (e.g.~a
word choice error that nevertheless makes sense in context), many errors can be detected this way
through judgments about coherence and understanding of the dialogue flow. Having
information about perceived mistakes could also be important for identifying those mistakes that go
unperceived in translation.

\begin{figure}[h!]
\centering
\includegraphics[width=0.8\linewidth]{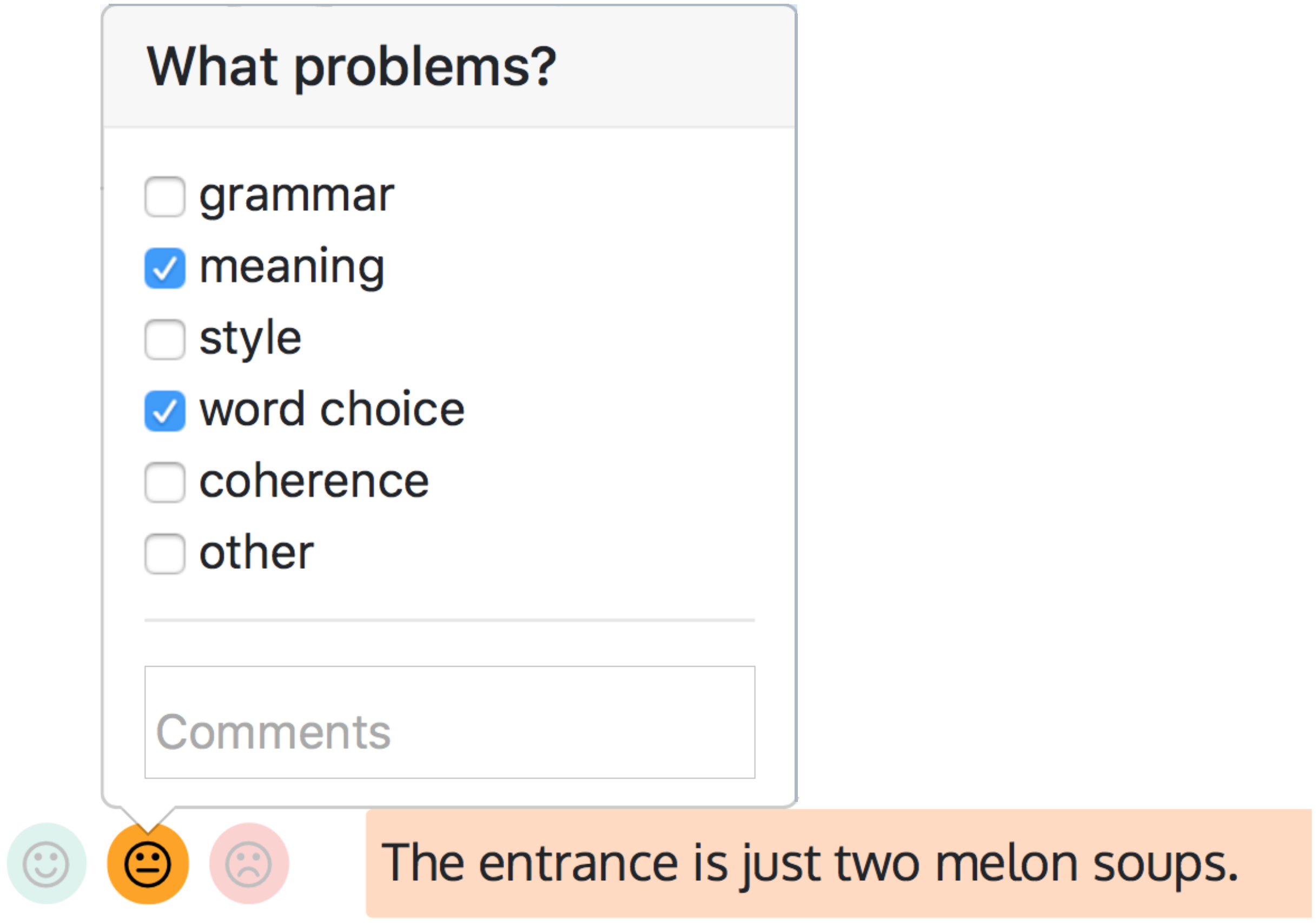}
\caption{\label{fig:eval} The sentence-level evaluation form. The original French sentence was
  \textit{L'entrée c'est juste deux soupes de melon.} ``The starter is just two melon soups.''}
\end{figure}

MT quality is evaluated twice, (i)~during the dialogue and (ii)~at the end of the
dialogue. Evaluations are saved for later and not shown to the other participant. They
evaluate each translated sentence during the dialogue by selecting an overall translation quality
(\textit{perfect}, \textit{medium} or \textit{poor}), and indicating which errors occur:
\textit{grammar}, \textit{meaning}, \textit{style}, \textit{word choice}, \textit{coherence} and
\textit{other} (see Fig.~\ref{fig:eval} for an example). Note that several problems can be indicated
  for the same sentence. If they wish, they can also write a free comment providing additional
information or suggesting corrections. Once the dialogue is finished, they give overall feedback of
the MT quality. They are asked to rank the quality of the translations in terms of
\textit{grammaticality}, \textit{meaning}, \textit{style}, \textit{word choice} and
\textit{coherence} on a five-point scale (\textit{excellent}, \textit{good}, \textit{average},
\textit{poor} and \textit{very poor}), and to indicate whether any particular aspects of
the translation or of the interface were problematic. Finally, they indicate whether they would
use such a system to communicate with a speaker of another language.

Participants were given instructions (with examples) on how to evaluate MT quality. However there is
expected to be a certain degree of variation in the way participants evaluate. This subjectivity,
inevitable with any human evaluation, is interesting, as it gives an indication of the variance of
the tolerance for errors, and which types of errors were considered most detrimental.


\paragraph{MT systems}
We compare the quality of two MT model types (see Sec.~\ref{sec:eval}). Within a dialogue, the
same model type is used for both language directions, and the same number of dialogues
is mediated by each model type. Both models are neural encoder-decoder models with attention
\citep{bahdanau_neural_2015}, implemented using Nematus \citep{sennrich_nematus:_2017}.
The first model (\textsc{baseline}) is trained to translate sentences in isolation.
The second (\textsc{2to2}), is trained to translate sentences in the context of the previous
sentence, as in \citep{tiedemann_neural_2017} and \citep{bawden_evaluating_2018}. This is done by
concatenating each sentence with its previous sentence, separated by a special token, and
translating both sentences at once. In a post-processing step, only the current sentence is
kept. Note that if the previous sentence is spoken by the same speaker as the current sentence,
then the original previous sentence is prepended. If the previous sentence is spoken by the other
speaker (in the opposite language), then the MT output of the previous sentence is prepended to the
current sentence. This means that the previous sentence is always in the same language as the
current sentence, and also corresponds to the context seen by the current speaker.

\paragraph{Training Data and MT setup}
The systems are trained using the OpenSubtitles2016 parallel corpus
\citep{lison_opensubtitles2016:_2016}. The data is cleaned, tokenised and truecased using the Moses
toolkit \citep{koehn_moses:_2007} and tokens are split into subword units using BPE
\citep{sennrich_neural_2016}. The data is then filtered to exclude poorly aligned or truncated sentences,
resulting in a training set of 24,140,225 sentences. Hyper-parameters are given in
App.~\ref{A:hyper}. During the dialogues, the participants' text is first split into sentences
and preprocessed in the same way as the training data. Translation is performed using
\textsc{Marian} for fast CPU decoding \citep{junczys-dowmunt_marian:_2018}. All processing scripts
will be released should the paper be accepted.

\section{Corpus characteristics}\label{sec:corpus}



\begin{table}[ht!]
\small
\centering
\scalebox{0.90}{
        
    \begin{tabular}{lrrr}
    \toprule
    Language direction& EN$\rightarrow$FR & FR$\rightarrow$EN & All  \\
    \midrule         
    \multicolumn{4}{c}{\#Turns}  \\ 
\midrule 
Total & 1,067 & 1,089 & 2,156 \\ 
Mean per dialogue & 7.4 & 7.6 & 15.0 \\ 
\midrule 
\multicolumn{4}{c}{\#Sentences}  \\ 
\midrule 
Total & 2,865 & 2,883 & 5,748 \\ 
Mean per dialogue & 19.9 & 20.0 & 39.9 \\ 
Min. / max. per dialogue & 5 / 42 & 5 / 60 & 10 / 102 \\ 
Mean per turn & 2.7 & 2.6 & 2.7 \\ 
Min. / max. per turn & 1 / 9 & 1 / 10 & 1 / 10 \\ 
\midrule 
\multicolumn{4}{c}{\#Tokens (original messages)}  \\ 
\midrule 
Total & 27,817 & 29,058 & 56,875 \\ 
Total unique & 3,588 & 4,244 & - \\ 
Mean per dialogue & 193.2 & 201.8 & 395.0 \\ 
Mean per sentence & 9.7 & 10.1 & 9.9 \\ 
\midrule 
\multicolumn{4}{c}{\#Tokens (MT versions)}  \\ 
\midrule 
Total & 28,745 & 27,510 & 56,255 \\ 
Total unique & 3,698 & 3,141 & - \\ 
Mean per dialogue & 199.6 & 191.0 & 390.7 \\ 
Mean per sentence & 10.0 & 9.5 & 9.8 \\ 
  \midrule 
\multicolumn{4}{c}{\#Tokens (reference translations)}  \\ 
\midrule 
Total & 30,093 & 27,014 & 57,107 \\ 
Total unique & 4,361 & 3,556 & - \\ 
Mean per dialogue & 209.0 & 187.6 & 396.6 \\ 
Mean per sentence & 10.5 & 9.4 & 9.9 \\ 
    \bottomrule      
    \end{tabular}}
\caption{\label{tab:dialogues} Corpus characteristics.}
\end{table}

Tab.~\ref{tab:dialogues} shows the basic characteristics of the
\totalNbDialogues{} dialogues. \percentDialoguesMoreThanThirtyfive{}\% of dialogues contain $\geq$35 sentences
and the average sentence length is 9.9 tokens, very slightly longer than the
translations. 
An extract of dialogue, representing a fictional argument, is given in Fig.~\ref{fig:dialogue},
providing an example of the type of language used by the participants. The language used is
colloquial and contains a number of fixed expressions (e.g.~\textit{get off your intellectual
  high-horse}, \textit{Mr Fancy pants}), which can prove difficult for MT, as is the case in this
example. The systems are sometimes robust enough to handle spelling and grammatical errors
(e.g.~\textit{qui ne penses} `who think$_{2.sg}$' instead of \textit{qui ne pense} `who
thinks$_{3.sg}$'' and \textit{rality} instead of \textit{reality}, translated into French as
\textit{ralité} instead of \textit{réalité}, conserving the spelling error in translation). The
dialogues also contain cultural references, such as references to films and actors. In many cases
named entities are well conserved, although sometimes cause problems (e.g.~\textit{Marcel Carné}
translated as \textit{Marcel Carborn}).\footnote{\textit{Carné} is segmented into two subwords,
  \textit{Car} and \textit{né} `born', and the second element has been translated.}



\begin{figure}[ht!]
\begin{mdframed}
\begin{itemize}[leftmargin=4.2mm,rightmargin=-1.5mm,itemsep=3pt,parsep=0pt]
\small
\item[EN:] Why do get the impression this is the first time you've mentioned anything going wrong between us?\\
\textcolor{gray}{\textit{MT:} Pourquoi pensez-vous que c'est la première fois que vous dites que quelque chose va mal entre
 nous~?} \\
Eval: \perfect
\item[FR:] Je suis comme Louis Jouvet dans les enfants du Paradis, quand il dit à Arletty qu'il a besoin de
changer d'air.\\
\textcolor{gray}{\textit{MT:} I'm like Louis Jouvet in the children of Heaven, when he tells Arletty
  he needs a change of air.} \\
\textit{Eval:} \perfect
\item[FR:] Tu te rappelles ce film magnifique de \textbf{Marcel Carné}, nous l'avons vu en français\\
\textcolor{gray}{\textit{MT:} You remember that beautiful \textbf{Marcel Carborn}, we saw it in
  French.} \\
\textit{Eval:} \medium (meaning) ```that''... could probably do without it'
\item[EN:] i have no idea what you're talking about.\\
\textcolor{gray}{\textit{MT:} Je ne sais pas de quoi tu parles.} \\
\textit{Eval:} \perfect
\item[EN:] But, i'm going to go with you on this one and assume that it has just slipped my mind.\\
\textcolor{gray}{\textit{MT:} Mais je vais y aller avec toi sur ce coup-là et assumer que ça m'a
  échappé.} \\
\textit{Eval:} \perfect
\item[EN:] so, yeah, sure, i remember your stupid film.\\
\textcolor{gray}{\textit{MT:} Donc, oui, bien sûr, je me souviens de ton stupide film.} \\
\textit{Eval:} \perfect
\item[EN:] it probably sucked.\\
\textcolor{gray}{\textit{MT:} C'était probablement naze.} \\
\textit{Eval:} \perfect
\item[] \dots
\item[FR:] J'ai besoin de stimulation intellectuelle, émotionnelle, pas de vivre à côté d'un geek attardé et
boutonneux qui ne \textbf{penses} que à ses jeux vidéos et à ses mangas.\\
\textcolor{gray}{\textit{MT:} I need intellectual stimulation, emotional stimulation, not to live next to a retarded and pimply-
faced geek who \textbf{thinks} only of his video games and its Manga.} \\
\textit{Eval:} \perfect
\item[EN:] well look at you, \textbf{Mr Fancy pants}! \\
\textcolor{gray}{\textit{MT:} Eh bien, regardez-moi ça, Mr.} \\
\textit{Eval:} \medium `the Mr isn't useful' 
\item[] \dots
\item[EN:] how about you get off your \textbf{intellectual high-horse} and get back to \textbf{rality} for a bit\\
\textcolor{gray}{\textit{MT:} Que diriez-vous de sortir de votre cheval intellectuel et de retourner à la \textbf{ralité} un
  moment~?} \\
\textit{Eval:} \medium (meaning) `cheval intellectuel doesn't exists'
\end{itemize}

\end{mdframed}
\caption{\label{fig:dialogue} A dialogue extract with baseline MT output (``MT:'', in grey) and
  human evaluation (``Eval:'').}
\end{figure}



\subsection{Normalised versions}

Although participants were encouraged to use their best spelling and grammar, such errors did occur
(missing or repeated words, typographical errors, inconsistent use of punctuation). We provide
manually normalised versions of sentences containing errors. The aim of this
normalisation is to provide information about the presence of errors (useful for
studying their impact on translation), and for providing a basis for the human
reference translations, as we do not attempt to reproduce errors in the
  translations. 
Corrections are kept to a minimum (i.e~non-canonical use of language was not corrected if linked to
the colloquial use of language), and therefore in practice are limited to the addition of capital letters
at the beginning of sentences and fullstops at the end of sentences and typographical error
correction only when the correct form can easily be guessed from the context.\footnote{When
  a wrong but attested word is used, but it is made explicit that this is intentional, we do
  not correct the word. E.g. the use of `deserts' instead of `desserts' in `The
  \textbf{deserts} are in the refrigerator. I said deserts for fun, I meant desserts!'}

\begin{table*}
\small
\centering
\scalebox{0.93}{
\begin{tabular}{lll}
\toprule
& Original utterance & Reference Translation \\
\midrule
Informal  & Well look at you, \textbf{Mr Fancy pants!} & Eh bien, regarde-toi,
                                                                  \textbf{M. le snobinard} ! \\
\cmidrule{1-3}
Self-correction & \ldots to do \textbf{uor} jobs\ldots & \ldots pour faire \textbf{notre} travail\ldots\\
 & Typo: \ldots to do \textbf{our} jobs\ldots &
Typo: \ldots pour faire \textbf{notre} travail\ldots \\
\cmidrule{1-3}
Ambiguity & D'ailleurs il est l'heure de \textbf{mon patient}$_{\text{male}}$ suivant. & Besides, it’s time
  for my next \textbf{patient}. \\
& Ou plutôt, de \textbf{ma patiente}$_{\text{female}}$ suivante, d'ailleurs. & Or \textbf{patient}
                                                                               should I say. \\
\cmidrule{2-3} 
& I can't stop thinking about ice-cream\ldots & Je ne peux pas m’empêcher d’avoir envie d’une glace\ldots\\ 
& \ldots à la \textbf{glace} qui se mange~? & \ldots about \textbf{ice-cream} that’s eaten?  \\
& ou bien à une \textbf{glace} pour se regarder ? & Or about a \textbf{mirror} to look into? \\
\cmidrule{1-3}
Meta-discussion & Tu connais un restau indonésien? & Do you know an
                                                                            Indonesian restaurant?
  \\
 & Ou \textbf{à la limite thaï}~? & Or at a push Thai? \textit{(MT: Or \textbf{the Thai limit})}\\
& What do you mean by \textbf{the Thai limit?} & Qu'est-ce que tu veux dire par \textbf{la limite thaïlandaise}~? \\
\bottomrule
\end{tabular}}
\caption{\label{tab:trad-probs} Examples of the three types of translation difficulties identified in 
\nameref{para:probs} (Sec.~\ref{para:probs})}
\end{table*}

\subsection{Machine translations}

Each sentence is translated automatically into the other language for the other participant. A
single type of MT system (\textsc{baseline} or \textsc{2to2}) is used for all sentences within a
dialogue. 
The use of two different systems is relevant to our
analysis of the human evaluations produced by dialogue participants (Sec.~\ref{sec:eval}). 
The choice of MT system does of course affect the quality of the MT. However, the corpus will remain
relevant and useful as a test set and for analysing human language behaviour in this setup
independently of this choice.

\subsection{Human reference translations}

In order for the corpus to be used as a test set for future MT models, we also produce human
reference translations for each language direction. Translators were native speakers of the target
language, with very good to bilingual command of the source language, and all translations were
further verified by a bilingual speaker.

Particular attention was paid to producing natural, spontaneous translations. The translators did
not have access to the machine translated versions of the sentences they were translating to
avoid any bias towards the MT models or the training data. However, they could see the machine
translated sentences of the opposite language direction. This was important to ensure that
utterances were manually translated in the context in which they were originally produced (as the
speaker would have seen the dialogue) and to ensure cohesive translations (e.g.~for discursive
phenomena, such as anaphoric phenomena and lexical repetition).
Spelling mistakes and other typographical irregularities (e.g.~missing punctuation and capital
letters) were not transferred to the translations;\footnote{Reproducing the same error type when
  translating is not always possible and so could depend on an arbitrary decision.} the
translations are therefore clean (as if no typographical errors had been present).

\paragraph{Translation difficulties}\label{para:probs}
The particularity of the setup makes a small number sentences difficult to translate (cf.~also the examples in Tab.~\ref{tab:trad-probs}):
\begin{itemize}
\item \textbf{the informal nature} means that many utterances are idiomatic and
  translation equivalents hard to find. We chose idiomatic
  equivalents based on communicative intention.
\item \textbf{ambiguity} in one language (but not the other) can sometimes lead to seemingly
  non-sensical utterances. This is a more theoretical translation problem and not one that can be
  solved satisfactorily. We translated these utterances as best as possible, despite resulting
  incoherences in the target language.
\item occasional coherence problems due to mistranslations (\textbf{a
    side-effect of using MT}), which led to meta-discussions. Where possible, we
made translations coherent with the dialogue context available to the speaker.
\end{itemize}


\subsection{Human judgments of MT quality}\label{sec:eval}

As described in Sec.~\ref{sec:protocol}, participants evaluated the translations from a
monolingual (target language) perspective. We provide a preliminary analysis of these judgments to
show that they are a good indicator of MT quality and that such a protocol is useful for
comparing MT models in a real setting.



\paragraph{Overall MT quality}\label{sec:mtquality}
Although an in-depth linguistic analysis is beyond the scope of this paper, we look here at global
trends in evaluation.\footnote{Given that the evaluation is perfomed by different participants on
  different subsets of sentences, comparisons should only be made when strong tendencies occur. The
  evaluations nevertheless provide a rich sort of information about how MT mistakes are perceived by
  different native speakers.} Differences between models are shown in
Fig.~\ref{fig:eval-comparative}. They show unsurprisingly that MT quality is dependent on the
language pair; translation into English is perceived as better than into French, approximately half
of all EN$\rightarrow$FR sentences being annotated as \textit{medium} or
\textit{poor}.\footnote{Note that these categories are used when any kind of problem is noticed in
  the sentence, however minor and whatever the nature of the error.}  There is little difference in
perceived quality between the \textsc{baseline} and \textsc{2to2} for FR$\rightarrow$EN. This
contrasts with EN$\rightarrow$FR, for which the number of sentences marked as \textit{perfect} is
higher by +4\% for \textsc{2to2} than for \textsc{baseline}. An automatic evaluation with BLEU
\footnote{Calculated using Moses'\texttt{multi-bleu-detok.perl}.}  \citep{papineni_bleu:_2002} shows
that the contextual model scores mildly better than the baseline, particularly for
EN$\rightarrow$FR. We retranslate all sentences with both models and compare the outputs to the
reference translations: for FR$\rightarrow$EN, the \textsc{2to2} model scores 31.34
(compared to 31.15 for the \textsc{baseline}), and for EN$\rightarrow$FR, \textsc{2to2} scores 
 31.60 (compared to 30.99 for the \textsc{baseline}).


\begin{figure}[ht!]
\begin{minipage}{\linewidth}
 \centering
 \small
\begin{tikzpicture}
\begin{axis}[
  name=enfr,
  title=English to French,
  title style={at={(0.5,1.15)},anchor=north},
  width=0.6\linewidth,
  height=4.5cm,
  x tick label style={ /pgf/number format/1000 sep=},
  ylabel=Percentage of sentences,
  ymajorgrids,
  enlarge x limits={abs=2.5em},
  label style={font=\small}, 
  tick label style={font=\small},
  legend style={at={(1,-0.2)}, anchor=north,legend columns=-1,draw=none},
  ybar=2pt,
  bar width=9pt,
  xtick=data,
  ymin=0,ymax=80,
  symbolic x coords={baseline,2-to-2},
  xticklabels={\textsc{baseline}, \textsc{2-to-2}},
  nodes near coords, 
  every node near coord/.append style={font=\tiny,xshift=+2.5pt}
]
\addplot[fill=brown!30!green] coordinates {(baseline,57.254623)(2-to-2,61.554333)};

\addplot[fill=white!10!orange] coordinates {(baseline,34.210526)(2-to-2,30.536451)};

\addplot[fill=red] coordinates {(baseline,8.534851)(2-to-2,7.909216)};

\legend{\textit{perfect}, \textit{medium}, \textit{poor}}
\end{axis}

\begin{axis}[
  at={($(enfr.south east)+(+0.5cm,+0mm)$)},
  title=French to English,
  title style={at={(0.5,1.15)},anchor=north},
  width=0.6\linewidth, 
  height=4.5cm,
  x tick label style={ /pgf/number format/1000 sep=},
  ymin=0,ymax=80,
  ymajorgrids,
  enlarge x limits={abs=2.5em},
  label style={font=\small},
  tick label style={font=\small}, 
  yticklabels={,,},
  ybar=2pt,
  bar width=9pt,
  xtick=data,
  symbolic x coords={baseline,2-to-2},
  xticklabels={\textsc{baseline}, \textsc{2-to-2}},
  nodes near coords,
  every node near coord/.append style={font=\tiny, xshift=+2.5pt}
]
\addplot[fill=brown!30!green] coordinates {(baseline,72.180959)(2-to-2,71.224052)};

\addplot[fill=white!10!orange] coordinates {(baseline,22.282242)(2-to-2,24.552613)};

\addplot[fill=red] coordinates {(baseline,5.536799)(2-to-2,4.223336)};

\end{axis}
\end{tikzpicture}
\caption{\label{fig:eval-comparative} Percentage of sentences for each language direction and model type
  marked as \textit{perfect}/\textit{medium}/\textit{poor} by participants.}
\vspace{0.6cm}
\end{minipage}
%
\begin{minipage}{\linewidth}
\centering
\small
\begin{tikzpicture}
\begin{axis}[
  name=enfr,
  title=\textsc{Baseline} English to French,
  ybar stacked,
  width=0.60\linewidth,
  height=4.2cm,
  ylabel=Percentage of sentences,
  label style={font=\small}, 
  tick label style={font=\small},
  x tick label style={rotate=35,anchor=east, xshift=+0.15cm,yshift=-0.15cm},     
  legend style={at={(1.95,0.96)},anchor=north,draw=none},
  legend cell align={left},   
  bar width=9pt,
  ymajorgrids,    
  xtick=data,
  ymin=0,ymax=20,
  symbolic x coords={grammar, style, word choice, meaning, coherence, other},
  xticklabels={\textit{grammar}\hphantom{o},
  \textit{style}\hphantom{o}, \textit{word choice}\hphantom{o}, 
  \textit{meaning}\hphantom{o}, \textit{coherence}, \textit{other}\hphantom{o}},     
]
\addplot[fill=white!20!orange] 
coordinates {(grammar,6.756757) (style,6.756757)(word choice,13.086771)
(meaning,3.627312)(coherence,5.761024)(other,0.2133711)};

\addplot[fill=red] 
coordinates {(grammar,1.066856) (style,0.995733)(word choice,2.773826)
(meaning,3.911807)(coherence,3.058321)(other,0)};

\legend{\textit{medium}, \textit{poor}}
\end{axis}
\begin{axis}[
  name=enfr2,
  at={($(enfr.south east)+(+0.5cm,0cm)$)},
  title=\textsc{2-to-2} English to French,
  stack plots=y,/tikz/ybar,    
  width=0.60\linewidth,
  height=4.2cm,
  x tick label style={rotate=35,anchor=east, xshift=+0.15cm,yshift=-0.15cm},   
  label style={font=\small}, 
  tick label style={font=\small},
  yticklabels={,,},
  ymajorgrids,
  bar width=9pt,
  xtick=data,
  ymin=0,ymax=20,
  symbolic x coords={grammar, style, word choice, meaning, coherence, other},
  xticklabels={\textit{grammar}\hphantom{o},
  \textit{style}\hphantom{o}, \textit{word choice}\hphantom{o}, 
  \textit{meaning}\hphantom{o}, \textit{coherence}, \textit{other}\hphantom{o}},   
]
\addplot[fill=white!20!orange] 
coordinates {(grammar,5.708391) (style,4.539202)(word choice,12.654746)
(meaning,3.026135)(coherence,3.920220)(other,0.343879)};

\addplot[fill=red] 
coordinates {(grammar,1.100413) (style,1.100413)(word choice,3.782669)
(meaning,2.957359)(coherence,1.994498)(other,0.206327)};
 \end{axis}
\begin{axis}[
  name=fren,
  at={($(enfr.south west)+(+0cm,-4.8cm)$)},
  title=\textsc{Baseline} French to English,
  ylabel=Percentage of sentences, 
  stack plots=y,/tikz/ybar,    
  width=0.60\linewidth,
  height=4.2cm,
  x tick label style={rotate=35,anchor=east, xshift=+0.15cm,yshift=-0.15cm},   
  label style={font=\small}, 
  tick label style={font=\small},
  bar width=9pt,
  ymajorgrids,
  xtick=data,
  ymin=0,ymax=20,
  symbolic x coords={grammar, style, word choice, meaning, coherence, other},
  xticklabels={\textit{grammar}\hphantom{o},
  \textit{style}\hphantom{o}, \textit{word choice}\hphantom{o}, 
  \textit{meaning}\hphantom{o}, \textit{coherence}, \textit{other}\hphantom{o}},    
]
\addplot[fill=white!20!orange] 
coordinates {(grammar,5.806887) (style,2.093180)(word choice,8.102633)
(meaning,4.118839)(coherence,1.890614)(other,0.540176)};

\addplot[fill=red] 
coordinates {(grammar,0.675219) (style,0.270088)(word choice,1.417961)
(meaning,2.093180)(coherence,1.282917)(other,0.067522)};

\end{axis}
\begin{axis}[
  name=fren2,
  at={($(fren.south east)+(+0.5cm,0cm)$)},
  title=\textsc{2-to-2} French to English,
  stack plots=y,/tikz/ybar,    
  width=0.60\linewidth,
  height=4.2cm,
  x tick label style={rotate=35,anchor=east, xshift=+0.15cm,yshift=-0.15cm},   
  label style={font=\small}, 
  tick label style={font=\small},
  bar width=9pt,
  yticklabels={,,},
  ymajorgrids,
  xtick=data,
  ymin=0,ymax=20,
  symbolic x coords={grammar, style, word choice, meaning, coherence, other},
  xticklabels={\textit{grammar}\hphantom{o},
  \textit{style}\hphantom{o}, \textit{word choice}\hphantom{o}, 
  \textit{meaning}\hphantom{o}, \textit{coherence}, \textit{other}\hphantom{o}},   
]
\addplot[fill=white!20!orange] 
coordinates {(grammar,5.654975) (style,2.505369)(word choice,9.735147)
(meaning,3.865426)(coherence,2.433787)(other,0.429492)};

\addplot[fill=red] 
coordinates {(grammar,0.715820) (style,0.214746)(word choice,1.073729)
(meaning,1.932713)(coherence,1.288475)(other,0.357910)};

\end{axis}
\end{tikzpicture}
\caption{\label{fig:eval_problems} Percentage of all sentences for each language direction and model
  type marked as containing each problem type (a sentence can have several problems). Bars
  are cumulative and show the percentage for sentences marked as \textit{medium} (orange) and
  \textit{poor} (red).}
\end{minipage}
\end{figure}

\paragraph{Types of errors encountered}

The evaluation results for each problem type are shown in Fig.~\ref{fig:eval_problems}. The
few number of problems classed as \textit{other} indicates that our categorisation of MT errors was
sufficiently well chosen. The most salient errors for all language directions and models are in
\textit{word choice}, especially when translating into French, with approximately 16\% of sentences
deemed to contain a word choice error. As with the overall evaluations, there are few differences
between \textsc{baseline} and \textsc{2to2} for FR$\rightarrow$EN, but significant differences are
seen for EN$\rightarrow$FR: \textsc{2to2} models perform better, with fewer errors in most problem
types, except \textit{word choice}. A notable difference is the lower frequency of
\textit{coherence}-related errors for \textsc{2to2}.  Coherence errors also appear to be less
serious, as there is a lower percentage of translations labelled as \textit{poor} (as opposed to
\textit{medium}).
These results are encouraging, as they show that our data collection method is a viable way to
collect human judgments, and that such judgments can reveal fine-grained differences in MT systems,
even when evaluating on different sentence sets. 

In spite of the errors, the translation quality is in general good, especially into English, and participant feedback is
excellent concerning intelligiblity and dialogue flow. As well as sentence-level
judgments, participants indicated overall MT quality once the dialogue was
complete. Participants indicated that they would use such a system to communicate with a speaker of
another language 89\% of the time. In 81\% of dialogues, grammaticality was marked as either
\textit{good} or \textit{excellent}. Coherence, style and meaning were all indicated as being
\textit{good} or \textit{excellent} between 76\% and 79\% of the time. As a confirmation of the
sentence-level evaluations, word choice was the most problematic error type, indicated in
only 56\% of dialogues as being \textit{good} or \textit{excellent} (40\% of dialogues had
\textit{average} word choice, leaving a very small percentage in which it was perceived as \textit{poor}).


\subsection{Focus on a discourse-level phenomenon}\label{sec:tuvous}

We study one specific discursive phenomenon: the
consistent use of French pronouns \textit{tu} and \textit{vous}. 
%
%
%
The French translation of singular \textit{you} is ambiguous between \textit{tu} (informal) and
\textit{vous} (formal). Their inconsistent use was one of the most commented problems by French
speakers,\footnote{Most of the scenarios were such that we would expect the same pronoun to be used
  by each speaker.} and a strategy for controlling this choice has been suggested for this reason
\citep{sennrich_controlling_2016}.  Neither of our models explicitly handles this choice, although
\textsc{2to2} does take into account pairs of consecutive sentences, and therefore could
be expected to have more consistent use of the pronouns across neighbouring sentences. As a proxy
for its ability to account for lexical cohesion,
we look at the two models' ability to ensure consistent translation of the pronouns across consecutive
sentences. For each model, we take translated sentences in which \textit{tu} or \textit{vous}
appear, and for which the previous sentence also contains either \textit{tu} or \textit{vous}. By
comparing the number of times the current sentence contains the same pronoun as the previous
sentence (see Tab.~\ref{tab:tuvous}), we can estimate the degree of translation consistency for
this particular aspect. Although the absolute figures are too low to provide statistical
significance, we can see a general trend that the \textsc{2to2} model does show greater consistency in
the use of the pronouns over the baseline model, with +\percentDiffTuTu{}\% in the consistency use of
\textit{tu} and +\percentDiffVousVous{}\% in the consistent use of \textit{vous}.


\begin{table}
\small
\centering
        
    \begin{tabular}{lrrrrrr}
    \toprule
    & \multicolumn{2}{c}{\textsc{Baseline}} && \multicolumn{2}{c}{\textsc{2to2}}   \\
    \multicolumn{1}{c}{Prev. $\backslash$ Curr.} & \multicolumn{1}{c}{\textit{tu}} & \multicolumn{1}{c}{\textit{vous}}
    && \multicolumn{1}{c}{\textit{tu}} & \multicolumn{1}{c}{\textit{vous}}\\
    \midrule         
    \textit{tu} & 49 & 31 &  & 55 & 23 &  \\ 
\textit{vous} & 27 & 27 &  & 22 & 28 &  \\ 
      
    \bottomrule      
    \end{tabular}
\caption{\label{tab:tuvous} For each model, the number of times each model translates using \textit{tu}
  and \textit{vous} and either of the forms \textit{tu} and \textit{vous} also appears in the previous sentence.}
\end{table}

\section{Discussion and future work}\label{sec:future}

Our new corpus of MT-mediated dialogues provides a basis for many future research studies, in terms
of the interaction between MT and humans: how good communication can be when using MT systems, how
MT systems must adapt to real-life human behaviour and how humans handle communication errors. We
have already shown through a preliminary analysis that the human judgments provide a viable form of
MT evaluation and can be further analysed to give us more insight into MT of dialogue. The same
protocol could be extended to other language pairs, although this would be part of a larger,
international effort.  

We intend to further extend the English-French corpus in future work and annotate it with
discourse-level information, which will pave the way for future phenomenon-specific evaluation:
how they are handled by different MT systems and evaluated by the participants.
In this direction, we manually annotated anaphoric phenomena in 27 dialogues (anaphoric
pronouns, event coreference, possessives, etc.). 
Despite the small size of this sample, it already displays interesting characteristics, which could provide
a strong basis for future work. Anaphoric references are common in the sample annotated: 250
anaphoric pronouns, 34 possessive pronouns, and 117 instances of event coreference. Their incorrect
translation was often a cause of communication problems (see
Fig.~\ref{fig:coref}), the impact of which will be investigated further.

\begin{figure}[ht!]
\begin{mdframed}
\small
FR$_{\mathrm{orig.}}$: Can I lie on this \textbf{couch}? \\
... \\
FR$_{\mathrm{trans.}}$: I don’t want to bother, \textbf{he} looks clean and new
\end{mdframed}
\caption{\label{fig:coref} An example of mistranslated coreference with the incorrect translation of
  French \textit{il} as \textit{he}, referring to \textit{canapé} `couch'.}
\end{figure}

The resource could also be extended to more scenarios. It is designed to cover a
wide range of topics, and to be relatively unrestricted in the scenarios imposed. This is a point of
difference with existing bilingual corpora, which are often task-orientated and/or limited to a
particular domain. For example BTEC (Basic Travel Expression Corpus) and SLDB (Spoken Language
DataBase) are both focused on the travel domain and represent a particular style of conversation
structured around a task. The relatively unrestricted nature of our corpus allows participants to be
freer in their language use, and will allow us to analyse conversational language over multiple
domains.

\section{Conclusion}
The corpus presented in this paper therefore provides a wide range of opportunities for future
research into the type of language used in this setting and the way in which participants handle
breakdowns. The protocol for data and MT judgment collection presented provides a useful framework
for future evaluation of MT quality, and the corpus itself can be used as a test set to guide manual
evaluation of new models. Our preliminary analyses of sentence-level human judgments show that the
evaluation procedure is viable, and we have observed some interesting differences between two types
of MT model used in our experiments.

\newpage
\bibliography{thesis}

\begin{thebibliography}{18}
\expandafter\ifx\csname natexlab\endcsname\relax\def\natexlab#1{#1}\fi

\bibitem[{Bahdanau et~al.(2015)Bahdanau, Cho, and
  Bengio}]{bahdanau_neural_2015}
Dzmitry Bahdanau, Kyunghyun Cho, and Yoshua Bengio. 2015.
\newblock Neural {Machine} {Translation} by {Jointly} {Learning} to {Align} and
  {Translate}.
\newblock In \emph{Proceedings of the 3rd {International} {Conference} on
  {Learning} {Representations}}, {ICLR}'15.
\newblock ArXiv: 1409.0473.

\bibitem[{Bawden et~al.(2018)Bawden, Sennrich, Birch, and
  Haddow}]{bawden_evaluating_2018}
Rachel Bawden, Rico Sennrich, Alexandra Birch, and Barry Haddow. 2018.
\newblock Evaluating {Discourse} {Phenomena} in {Neural} {Machine}
  {Translation}.
\newblock In \emph{Proceedings of the 2018 {Conference} of the {North}
  {American} {Chapter} of the {Association} for {Computational} {Linguistics};
  {Human} {Language} {Technologies}}, {NAACL}-{HLT}'18, pages 1304--1313, New
  Orleans, Louisiana, USA.

\bibitem[{Dyer et~al.(2013)Dyer, Chahuneau, and Smith}]{dyer_simple_2013}
Chris Dyer, Victor Chahuneau, and Noah~A. Smith. 2013.
\newblock A simple, fast, and effective reparameterization of {IBM} {Model} 2.
\newblock In \emph{Proceedings of the {Conference} of the {North} {American}
  {Chapter} of the {Association} for {Computational} {Linguistics}: {Human}
  {Language} {Technologies}}, {NAACL}-{HLT}'13, pages 644--648, Denver,
  Colorado, USA.

\bibitem[{Federmann and Lewis(2016)}]{federmann_microsoft_2016}
Christian Federmann and Will Lewis. 2016.
\newblock Microsoft {Speech} {Language} {Translation} ({MSLT}) {Corpus}: {The}
  {IWSLT} 2016 release for {English}, {French} and {German}.
\newblock \emph{Microsoft Research}.

\bibitem[{Higashinaka et~al.(2016)Higashinaka, Funakoshi, Kobayashi, and
  Inaba}]{Higashinaka_dialogue_2016}
Ryuichiro Higashinaka, Kotaro Funakoshi, Yuka Kobayashi, and Michimasa Inaba.
  2016.
\newblock The {Dialogue} {Breakdown} {Detection} {Challenge}: {Task}
  {Description}, {Datasets}, and {Evaluation} {Metrics}.
\newblock In \emph{Proceedings of the 10th {International} {Conference} on
  {Language} {Resources} and {Evaluation}}, {LREC}'16, pages 3146--3150,
  Portorož, Slovenia.

\bibitem[{Isabelle et~al.(2017)Isabelle, Cherry, and
  Foster}]{isabelle_challenge_2017}
Pierre Isabelle, Colin Cherry, and George Foster. 2017.
\newblock A {Challenge} {Set} {Approach} to {Evaluating} {Machine}
  {Translation}.
\newblock In \emph{Proceedings of the 2017 {Conference} on {Empirical}
  {Methods} in {Natural} {Language} {Processing}}, {EMNLP}'17, pages
  2476--2486, Copenhagen, Denmark.

\bibitem[{Junczys-Dowmunt et~al.(2018)Junczys-Dowmunt, Grundkiewicz, Dwojak,
  Hoang, Heafield, Neckermann, Seide, Germann, Aji, Bogoychev, Martins, and
  Birch}]{junczys-dowmunt_marian:_2018}
Marcin Junczys-Dowmunt, Roman Grundkiewicz, Tomasz Dwojak, Hieu Hoang, Kenneth
  Heafield, Tom Neckermann, Frank Seide, Ulrich Germann, Alham~Fikri Aji,
  Nikolay Bogoychev, André F.~T. Martins, and Alexandra Birch. 2018.
\newblock Marian: {Fast} {Neural} {Machine} {Translation} in {C}++.
\newblock \emph{arXiv:1804.00344 [cs]}.
\newblock ArXiv: 1804.00344.

\bibitem[{King and Falkedal(1990)}]{king_using_1990}
Margaret King and Kirsten Falkedal. 1990.
\newblock Using test suites in evaluation of machine translation systems.
\newblock In \emph{Proceedings of the 1990 {Conference} on {Computational}
  {Linguistics}}, {COLING}'90, pages 211--216, Helsinki, Finland.

\bibitem[{Koehn et~al.(2007)Koehn, Hoang, Birch, Callison-Burch, Federico,
  Bertoldi, Cowan, Shen, Moran, Zens, Dyer, Bojar, Constantin, and
  Herbst}]{koehn_moses:_2007}
Philipp Koehn, Hieu Hoang, Alexandra Birch, Chris Callison-Burch, Marcello
  Federico, Nicola Bertoldi, Brooke Cowan, Wade Shen, Christine Moran, Richard
  Zens, Chris Dyer, Ondřej Bojar, Alexandra Constantin, and Evan Herbst. 2007.
\newblock Moses: {Open} {Source} {Toolkit} for {Statistical} {Machine}
  {Translation}.
\newblock In \emph{Proceedings of the 45th {Annual} {Meeting} of the
  {Association} for {Computational} {Linguistics}}, {ACL}'07, pages 177--180,
  Prague, Czech Republic.

\bibitem[{Lison and Tiedemann(2016)}]{lison_opensubtitles2016:_2016}
Pierre Lison and Jörg Tiedemann. 2016.
\newblock {OpenSubtitles}2016: {Extracting} {Large} {Parallel} {Corpora} from
  {Movie} and {TV} {Subtitles}.
\newblock In \emph{Proceedings of the 10th {Language} {Resources} and
  {Evaluation} {Conference}}, {LREC}'16, pages 923--929, Portorož, Slovenia.

\bibitem[{Morimoto et~al.(1994)Morimoto, Uratani, Takezawa, Furuse, Sobashima,
  Iida, Nakamura, Sagisaka, Higuchi, and Yamazaki}]{morimoto_speech_1994}
Tsuyoshi Morimoto, Noriyoshi Uratani, Toshiyuki Takezawa, Osamu Furuse,
  Yasuhiro Sobashima, Hitoshi Iida, Atsushi Nakamura, Yoshinori Sagisaka, Norio
  Higuchi, and Yasuhiro Yamazaki. 1994.
\newblock A {Speech} and {Language} {Database} for {Speech} {Translation}
  {Research}.
\newblock In \emph{Proceedings of the 3rd {International} {Conference} on
  {Spoken} {Language} {Processing}}, {ICSLP}'94, pages 1791--1794, Yokohama,
  Japan.

\bibitem[{Papineni et~al.(2002)Papineni, Roukos, Ward, and
  Zhu}]{papineni_bleu:_2002}
Kishore Papineni, Salim Roukos, Todd Ward, and Wei-Jing Zhu. 2002.
\newblock {BLEU}: {A} {Method} for {Automatic} {Evaluation} of {Machine}
  {Translation}.
\newblock In \emph{Proceedings of the 40th {Annual} {Meeting} of the
  {Association} for {Computational} {Linguistics}}, {ACL}'02, pages 311--318,
  Philadelphia, Pennsylvania, USA.

\bibitem[{Sennrich et~al.(2017)Sennrich, Firat, Cho, Birch, Haddow, Hitschler,
  Junczys-Dowmunt, Läubli, Valerio, Barone, Mokry, and
  Nădejde}]{sennrich_nematus:_2017}
Rico Sennrich, Orhan Firat, Kyunghyun Cho, Alexandra Birch, Barry Haddow,
  Julian Hitschler, Marcin Junczys-Dowmunt, Samuel Läubli, Antonio Valerio,
  Miceli Barone, Jozef Mokry, and Maria Nădejde. 2017.
\newblock Nematus: a {Toolkit} for {Neural} {Machine} {Translation}.
\newblock In \emph{Proceedings of the {Software} {Demonstrations} of the 15th
  {Conference} of the {European} {Chapter} of the {Association} for
  {Computational} {Linguistics}}, {EACL}'17, pages 65--68, Valencia, Spain.

\bibitem[{Sennrich et~al.(2016{\natexlab{a}})Sennrich, Haddow, and
  Birch}]{sennrich_controlling_2016}
Rico Sennrich, Barry Haddow, and Alexandra Birch. 2016{\natexlab{a}}.
\newblock Controlling {Politeness} in {Neural} {Machine} {Translation} via
  {Side} {Constraints}.
\newblock In \emph{Proceedings of the 2016 {Conference} of the {North}
  {American} {Chapter} of the {Association} for {Computational} {Linguistics}:
  {Human} {Language} {Technologies}}, {NAACL}-{HLT}'16, pages 35--40, San
  Diego, California, USA.

\bibitem[{Sennrich et~al.(2016{\natexlab{b}})Sennrich, Haddow, and
  Birch}]{sennrich_neural_2016}
Rico Sennrich, Barry Haddow, and Alexandra Birch. 2016{\natexlab{b}}.
\newblock Neural {Machine} {Translation} of {Rare} {Words} with {Subword}
  {Units}.
\newblock In \emph{Proceedings of the 54th {Annual} {Meeting} of the
  {Association} for {Computational} {Linguistics}}, {ACL}'16, pages 1715--1725,
  Berlin, Germany.

\bibitem[{Takezawa et~al.(2007)Takezawa, Kikui, Mizushima, and
  Sumita}]{Takezawa_multilingual_2007}
Toshiyuki Takezawa, Genichiro Kikui, Masahide Mizushima, and Eiichiro Sumita.
  2007.
\newblock Multilingual {Spoken} {Language} {Corpus} {Development} for
  {Communication} {Research}.
\newblock \emph{Computational Linguistics and Chinese Language Processing},
  12(3):303--324.

\bibitem[{Takezawa et~al.(2002)Takezawa, Sumita, Sugaya, Yamamoto, and
  Yamamoto}]{takezawa_toward_2002}
Toshiyuki Takezawa, Eiichiro Sumita, Fumiaki Sugaya, Hirofumi Yamamoto, and
  Seiichi Yamamoto. 2002.
\newblock Toward a {Broad}-coverage {Bilingual} {Corpus} for {Speech}
  {Translation} of {Travel} {Conversations} in the {Real} {World}.
\newblock In \emph{Proceedings of the 3rd {International} {Conference} on
  {Language} {Resources} and {Evaluation}}, {LREC}'02, pages 147--152, Las
  Palmas, Canary Islands, Spain.

\bibitem[{Tiedemann and Scherrer(2017)}]{tiedemann_neural_2017}
Jörg Tiedemann and Yves Scherrer. 2017.
\newblock Neural {Machine} {Translation} with {Extended} {Context}.
\newblock In \emph{Proceedings of the 3rd {Workshop} on {Discourse} in
  {Machine} {Translation}}, {DISCOMT}'17, pages 82--92, Copenhagen, Denmark.

\end{thebibliography}
\bibliographystyle{acl_natbib}

\cleardoublepage
\appendix

\pagenumbering{gobble}

\appendix

\section*{Appendices}

\section{Scenarios}\label{A:scenarios}

The following are the English versions of the scenarios and speaker roles. These (and all
instructions) are presented in French to French speakers.

\begin{mdframed}
\small

\quad\textbf{You are both lost in a forest.}\\
Roles: N/A\\

\textbf{You are chefs preparing a meal.}\\
Role 1:
You are the head chef and you are talking to your subordinate.\\
 Role 2:
You are the subordinate chef and you are talking to the head chef.\\

\textbf{You are in a classroom.}\\
Role 1:
You are the teacher and you are talking to a student.\\
 Role 2:
You are the student and you are talking to your teacher.\\

\textbf{You are feeding the ducks by the pond.}\\
Roles: N/A\\

\textbf{You are both organising a party.}\\
Role 1:
It's your party.\\
 Role 2:
It's their party.\\

\textbf{You are both stuck in a lift at work.}\\
Role 1:
You are an employee and you are with your boss.\\
 Role 2:
You are the boss and are with an employee.\\

\textbf{You are in a retirement home.}\\
Role 1:
You are visiting and talking to an old friend.\\
 Role 2:
You are a resident and you are talking with an old friend who is visiting you.\\

\textbf{You are in a bar.}\\
Role 1:
You are the bartender and talking to a customer.\\
 Role 2:
You are a customer and are talking to the bartender.\\

\textbf{You are in an aeroplane.}\\
Role 1:
You are scared and are speaking to the person sitting next to you.\\
 Role 2:
You are speaking to the person next to you, who is scared.\\

\textbf{You are at home in the evening.}\\
Role 1:
You are telling your spouse about the awful day you had.\\
 Role 2:
You are listening to your spouse telling you about the awful day they had.\\

\textbf{You are in a psychiatrist's consulting room.}\\
Role 1:
You are the psychiatrist and are with your patient.\\
 Role 2:
You are a patient and you are talking to your psychiatrist.\\

\textbf{You are on holiday by the pool.}\\
Role 1:
You are trying to relax and the other person wants to do something else.\\
 Role 2:
You want to do something else and the other person is trying to relax.
\end{mdframed}

\section{MT Hyper-parameters}\label{A:hyper}

Hyper-parameters for both MT models:

\begin{itemize}
\item 90,000 BPE operations, and threshold = 50
\item Filtering out of parallel sentences in which fewer than 80\% of tokens are aligned (after
  running FastAlign \citep{dyer_simple_2013})
\item Embedding layer dimension = 512, hidden layer dimension = 1024, batch size = 80, tied decoder
embeddings and layer normalisation, maximum sentence length = 50
\end{itemize}

\section{Participation guidelines}

The following guidelines were presented to all participants, and were available during the dialogue
if needed. A French translation was presented to French-speaking participants. 

\subsection*{DiaBLa Instructions}

You will be participating in an improvised written dialogue with another user. You will each write
in your own native language (English or French). Don't worry - you do not need to be able to speak
or understand the other language. Machine translation systems will translate all of the other
person's sentences into your language. You will also evaluate the translations from a monolingual
point of view (i.e. is the sentence grammatical? Does it make sense? Was the word choice ok? Is it
stylistically appropriate? Is it coherent with respect to previous sentences?)

Please read all instructions carefully before continuing!

 
\subsection{Signing up and logging in}
You must first register (we require some basic information -- see FAQ).  Log in using the email address you
registered with. Choose a username and the language you are going to speak in, which must be your
mother tongue.  You will be talking to real people. To increase your chances of finding someone to
chat to, fill in this spreadsheet with your availabilities. Or try your luck and log in straight
away!

\subsection{Dialoguing}
Once logged in, invite someone by clicking on their username or wait for someone to invite you. You
can accept or refuse an invitation to dialogue. If the request is accepted, you will be taken to the
dialogue screen. One of you is assigned the first turn, and after that, you are free to dialogue as
you please.  You will be presented with a setting (at the top of the chat box) in which the dialogue
will take place. E.g.~``You are in a forest" and your role.  Now improvise a dialogue in the setting
provided, as in improvised drama or role-play. I.e.~play a made-up character and not yourself. The
dialogues are to be like written drama transcriptions, rather than chat messages. We recommend
writing at least 15 sentences each (you will receive a message when this happens). You can write
more, but it is even more useful for
us to have more dialogues rather than fewer very long ones.  \\

\frownie{} \textbf{Please do not use:}
\begin{itemize}
\item emoticons or SMS speech 
\item your partner's username, your own username or personal details 
\end{itemize}

\smiley{} \textbf{Please do use:}
\begin{itemize}
\item your best spelling, grammar and punctuation  
\item the correct gender of you and your partner (for instance when
using pronouns) 
\item your imagination! You can refer to imaginary objects/people around you 
\end{itemize} 

\subsection{Evaluation}
You will evaluate the other person's translated utterances by selecting one of the smileys: 
\begin{itemize}
\item green smiley face: ``perfect" 
\item orange neutral face: ``ok but not perfect" 
\item sad red face: ``poor" 
\end{itemize}

When you select a smiley, you will be prompted to indicate what is wrong with the translation:
grammar, meaning, word choice, style, coherence, plus any extra comments to make your evaluation
clearer. See FAQ for some examples.  

\subsection{Purpose}
We will be using the dialogues to evaluate the machine translation systems and how easy
communication was. The dialogues will be used to create a corpus of dialogues, which will be freely
distributed for research purposes, and also used to analyse the machine translation models. Be
natural, spontaneous and creative! However, please avoid making your sentences purposefully
difficult in order to trick the machine translation system. Thank you!

\begin{figure*}[hbt!]
\centering
\includegraphics[height=380px]{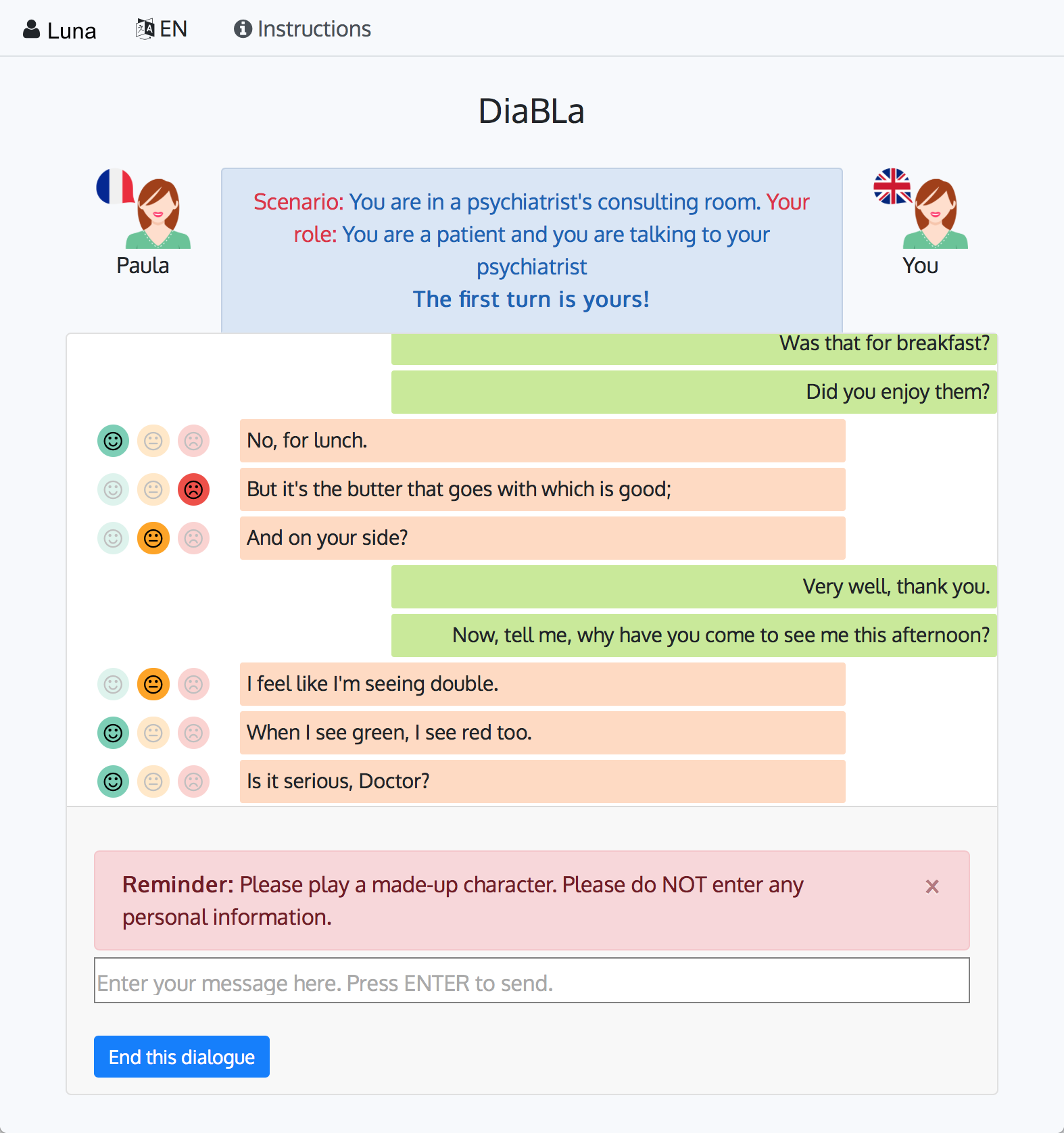}
\caption{\label{fig:interface} The dialogue interface for the English participant.}
\end{figure*}

\subsection{FAQ}

\textit{What if I don't understand what my partner says? }\\
As in real life, speak about the problem with your partner. Say that you don't understand and try to continue the dialogue as best as possible.\\

\textit{When evaluating, what do the error types correspond to?}
\begin{itemize}
\item Grammatical: the sentence is agrammatical. A few examples:
(i)~number disagreement: ``The boy are there.", (ii)~Missing articles: ``I want dog.", (iii)~Wrong
use of tenses, (iv)~Gender disagreement (for French), etc.
\item Meaning: the sentence does not appear to make sense, e.g.:
I was told by my avocado that a sentence was likely.
\item Word choice: a poor word choice was made, e.g.:~``I did you a chocolate cake" (instead of ``I made you a chocolate cake."), ``He took an attempt" (instead of ``He made an attempt")
\item Style: the level of formality is inconsistent or language usage is strange, although grammatically well-formed and understandable, e.g.:~Strange/unnatural utterances, wrong level of formality: ``What's up" in a job interview, etc.
\item Coherence: Lack of consistency with previous utterances or the context: wrong pronoun
  used that refers to something previously mentioned, inconsistent use of ``tu" and
  ``vous" (for French), word choice is inconsistent with what was previously said (e.g.~``I'm angry! – What do you
  mean by 'upset'?"), etc.
\end{itemize}

\textit{Why do you need personal information? }
\\
This enables us to evaluate whether certain aspects of the conversation (e.g. gender marking in
French) are correctly translated or not. It allows us to analyse how machine translation systems
react to the differences in language use, which depends for instance on the age of the user. The
personal information that will be distributed in the resulting corpus is the following:
\begin{itemize}
\item Your age bracket (18-24, 25-34, 35-44, etc.)  
\item Your gender
\item  Your French and English ability 
\item The other languages you speak 
\item Whether or not you have studied/worked in Natural Language
  Processing or research 
\end{itemize}

\textit{Why do you need to know speaker gender?}
\\
Speaker gender can be important in certain languages in terms of which words agree with the gender
of the speaker (e.g. French). We therefore ask you to choose between male and female for practical
reasons. If you do not identify with either gender, please choose the one by which you wish to be
identified linguistically (i.e. would you prefer to be referred to as ``he" or ``she"?). The important
thing is to be coherent when you dialogue in your use of gender.

\section{Dialogue interface}

Figure~\ref{fig:interface} shows an example of the interface used for dialogue collection.

\end{document}